\documentclass[nohyperref]{article}

\usepackage{microtype}
\usepackage{graphicx}
\usepackage{subfigure}
\usepackage{booktabs} 

\usepackage{hyperref}

\usepackage[accepted]{icml2023}

\usepackage{amsmath}
\usepackage{amssymb}
\usepackage{mathtools}
\usepackage{amsthm}
 \usepackage{multirow}
 \usepackage{caption}
 \usepackage{enumitem}
\usepackage[capitalize,noabbrev]{cleveref}

\theoremstyle{plain}

\theoremstyle{definition}

\theoremstyle{remark}

\usepackage[textsize=tiny]{todonotes}

\icmltitlerunning{Self-Supervised Learning of Split Invariant Equivariant Representations}

\begin{document}

\twocolumn[
\icmltitle{Self-Supervised Learning of Split Invariant Equivariant Representations}

\begin{icmlauthorlist}
\icmlauthor{Quentin Garrido}{meta,uge}
\icmlauthor{Laurent Najman}{uge}
\icmlauthor{Yann LeCun}{meta,nyu1,nyu2}
\end{icmlauthorlist}

\icmlaffiliation{meta}{Meta AI - FAIR}
\icmlaffiliation{uge}{Univ Gustave Eiffel, CNRS, LIGM, F-77454 Marne-la-Vallée, France}
\icmlaffiliation{nyu1}{Courant Institute, New York University}
\icmlaffiliation{nyu2}{Center for Data Science, New York University}

\icmlcorrespondingauthor{Quentin Garrido}{garridoq@meta.com}

\icmlkeywords{Machine Learning, self-supervised learning, equivariance, invariance, representation learning}

\vskip 0.3in
]

\printAffiliationsAndNotice{} 

\begin{abstract}
Recent progress has been made towards learning invariant or equivariant representations with self-supervised learning. While invariant methods are evaluated on large scale datasets, equivariant ones are evaluated in smaller, more controlled, settings. We aim at bridging the gap between the two in order to learn more diverse representations that are suitable for a wide range of tasks. We start by introducing a dataset called 3DIEBench, consisting of renderings from 3D models over  55 classes and more than 2.5 million images where we have full control on the transformations applied to the objects. We further introduce a predictor architecture based on hypernetworks to learn equivariant representations with no possible collapse to invariance. We introduce SIE (\textbf{S}plit \textbf{I}nvariant-\textbf{E}quivariant) which combines the hypernetwork-based predictor with representations split in two parts, one invariant, the other equivariant, to learn richer representations. We demonstrate significant performance gains over existing methods on equivariance related tasks from both a qualitative and quantitative point of view. We further analyze our introduced predictor and show how it steers the learned latent space. We hope that both our introduced dataset and approach will enable learning richer representations without supervision in more complex scenarios.
Code and data are available at \href{https://github.com/facebookresearch/SIE}{https://github.com/facebookresearch/SIE}.
\end{abstract}

\section{Introduction}

Self-supervised learning of image representations has made significant progress in recent years~\citep{chen2020simple,he2020moco, chen2020mocov2, grill2020byol, lee2021cbyol, caron2020swav, zbontar2021barlow, bardes2021vicreg, tomasev2022relicv2, caron2021dino, chen2021mocov3, li2022esvit, zhou2022ibot, zhou2022mugs,haochen2021provable,he2022masked, bardes2022vicregl}, catching up to supervised baselines in tasks requiring high-level information such as classification. Most of these works are placed in a joint-embedding framework, where two augmented views are generated from a source image. These two views are then fed to an encoder, giving \textit{representations}, and then through a projection head, giving \textit{embeddings}. Finally, a loss minimises the distance between the embeddings, i.e. makes them invariant to the augmentations, and is combined with a regularisation loss to spread embeddings in space.

While these invariance based approaches have been very successful for classification when using augmentations that preserve the semantic information of the image, this removal of information may be problematic for downstream tasks. For example, the use of color-jitter removes color information which can be useful for tasks such as flower classification~\citep{lee_improving_2021}. This motivates the goal of introducing equivariance to representations, in order to learn more general representations for more varied downstream tasks. We say that representations are equivariant if the application of data augmentation commutes with the application of the encoder, i.e. can the representations of two related views be mapped similarly as the views themselves.
Previous works have introduced ways to enrich usually invariant representations by keeping information about the augmentations. One approach is to use subsets of augmentations to construct partially invariant representations~\citep{xiao_what_2021}.
This can also be done by predicting rotations~\citep{dangovski_equivariant_2022}, preserving augmentation strengths in the representations~\citep{xie2022should}, or by predicting all of the augmentation parameters~\citep{lee_improving_2021} or a discretized version of them~\citep{scherr2022selfsupervised}.
While a mapping between representations may exist with these approaches, they offer no guarantees on its existence nor on its complexity. There are also no reliable approaches to empirically prove its existence. As such we do not consider these methods to truly be equivariant.
Learning equivariant representations requires being able to predict a representation from another in latent space, which has also been a successful paradigm. This can be done by a simple prediction head, either using reconstruction~\citep{winter2022unsupervised} or without~\citep{park_learning_2022,devillers2022equimod,shakerinava2022structuring}.

Parameter prediction based methods have been developed for image datasets such as ImageNet~\citep{deng2009imagenet} where there is no clear equivariant task and where augmentations happen in pixel space with no loss of information. On the other hand, equivariance based methods have been used on simpler synthetic datasets\cite{shakerinava2022structuring} where we can evaluate equivariance, but where it is hard to evaluate other classical computer vision tasks such as classification.
To bridge the gap between those two worlds, we start by introducing a dataset called 3DIEBench, consisting of renderings of over fifty-thousand 3D objects where we can study an equivariance related task (3D rotation prediction) and an invariant one (image classification). This allows us to measure more precisely how invariant classical self-supervised methods are, while also showing limitations of existing equivariant approaches where predictors often collapse to the identity, leading to invariant representations.

We then introduce a hypernetwork~\citep{ha2016hypernetworks} based predictor which avoids a collapse to the identity by design and show how it can outperform existing predictor architectures. We further show that by separating the representations in equivariant and invariant parts, we can significantly improve performance on equivariance related tasks, allowing us to match supervised baselines.
To complement our quantitative results we also analyze qualitatively the learned split invariant-equivariant representations and see that all invariant information is not discarded from the equivariant part, and that the predictor offers a meaningful way to steer the latent space.
To summarize:
\begin{itemize}[nosep]
    \item We introduce 3DIEBench, a new dataset to evaluate representations on tasks that require invariant and equivariant information
    \item We show the limitations of existing predictor architectures and introduce a hypernetwork based one that improves performance on all methods
    \item We show that splitting the representations in invariant and equivariant parts further improves performance on equivariance related tasks
\end{itemize}

\section{Related works}

\paragraph{Invariant Self-supervised learning}
Two main families of methods can be distinguished: contrastive and non-contrastive. Contrastive methods~\citep{chen2020simple,he2020moco,chen2020mocov2,chen2021mocov3,yeh2021decoupled} mostly rely on the InfoNCE criterion~\citep{oord2018infonce} except for~\cite{haochen2021provable} which uses squared similarities between the embedding. A clustering variant of contrastive learning has also emerged~\citep{caron2018clustering,caron2020swav,caron2021dino} and can be thought of as contrastive methods, but between cluster centroids instead of samples. Non-contrastive methods~\citep{grill2020byol,chen2020simsiam,bardes2021vicreg,zbontar2021barlow,ermolov2021whitening,li2022neural,bardes2022vicregl} aim at bringing together embeddings of positive samples, similar to contrastive learning. However, a key difference with contrastive methods lies in how those methods prevent a representational collapse. In the former, the criterion explicitly pushes away negative samples, i.e., all samples that are not positive, from each other. In the latter, the criterion considers the embeddings as a whole and encourages information content maximization to avoid collapse, e.g., by regularizing the empirical covariance matrix of the embeddings. While we study methods from both families in our experiments, they have been shown to lead to very similar representations~\cite{garrido2022duality}.

\paragraph{Introducing equivariance in invariant self-supervised learning}
While most of the aforementioned works focus on learning representations that are invariant to augmentations, some works have instead tried to learn representations where information about certain transformations is preserved. This can be done by predicting the augmentation parameters~\citep{scherr2022selfsupervised,lee_improving_2021,gidaris2018unsupervised}, or by introducing other transformations such as image rotations~\citep{dangovski_equivariant_2022}. Preserving the augmentations' strength in the representations can also be used to learn less invariant representations~ \citep{xie2022should}. 
These methods offer no guarantees on the existence of a mapping between transformed representations in latent space, nor ways to prove its existence or lack thereof. As such these methods cannot be considered to truly be equivariant.

\paragraph{Equivariant representation learning}
Previous works have explored equivariant representation learning using autoencoders, such as transforming autoencoders~\cite{hinton2011transforming}, Homeomorphic VAEs~\cite{falorsi2018explorations} or \citep{winter2022unsupervised}.
Recent works such as EquiMod~\citep{devillers2022equimod} or SEN~\citep{park_learning_2022} have also included a predictor that enables the steering of representations in latent space, without requiring reconstruction. These methods form the basis for our comparisons. In~\cite{marchetti2022equivariant}, representations are split in class and pose, i.e. invariant and equivariant, and assumes a simple equivariant latent space where the group action is the same as in the underlying data, e.g. 3 dimensions to represent pose. This assumes prior knowledge on the group of transformations, and can prove limited when the transformations cause a loss of information. Transformations are also assumed to be small, similarly as for SEN. We aim at deriving a more general predictor architecture with no such priors.
In~\cite{shakerinava2022structuring}, equivariant representations are learned with no knowledge of the group element associated with the transformation, but by having pairs of samples where the same transformation was applied.

\section{3DIEBench: A new benchmark for invariant-equivariant SSL}

Existing datasets used to evaluate equivariant or invariant representations have flaws when trying to design a method to learn more general representations. Datasets used to evaluate equivariance in representations often consist of simple images due to the need to control how transformations are applied~\cite{park_learning_2022,kipf2019contrastive}.
Conversely, datasets used to evaluate invariant representations~\cite{deng2009imagenet,cifar} are limited in the sense that position and shape of the objects in these dataset can not be parameterized by controllable transformations, and only pixel-level transformations can be applied on the images.
This motivates us to introduce a new dataset called 3D Invariant Equivariant Benchmark (3DIEBench) that aims at bridging the gap between the two.\\
We want a dataset that is not trivial for an invariant task (image classification) but where we still have control on the parameters of the scene and the objects within it to learn meaningful equivariant representations.
Taking inspiration from 3DIdent~\cite{zimmermann2021contrastive,von_kugelgen_self-supervised_2021} we use renderings of 3D objects from the subset of ShapeNetCore~\cite{chang2015shapenet} originating from 3d Warehouse~\cite{3dwarehouse}. This gives us a total 52472 objects spread across 55 classes. We are then able to adjust various factors of variations such as the object rotation, the lighting color, or the floor color. We focus on learning representations that are equivariant with respect to object rotations of arbitrary strength due to their inherent difficulty when using a diverse dataset, as well as the loss of information they can cause when looking at 2d renderings of the scene. In our experiments, we constrain the range of rotations to Euler angles between $-\frac{\pi}{2}$ and $\frac{\pi}{2}$. The goal is to make the task tractable, while still remaining challenging, as we show in our experiments. Using arbitrary rotations on arbitrary objects can make the task close to impossible, even for primates~\citep{logothetis1994view}.\\
For each object, we generate 50 random values for the factors of variation and then render the scene using Blender~\cite{blender} and BlenderProc~\cite{denninger2019blenderproc}, for a total of around 2.5 million images.
Sample renderings can be found in figure~\ref{fig:data_sample}. In supplementary section~\ref{sec:dataset} we provide more details on the dataset generation as well as additional visualizations. The dataset as well as the code to generate the renderings will be released.

\begin{figure}[!t]
    \centering
    \includegraphics[width=\columnwidth]{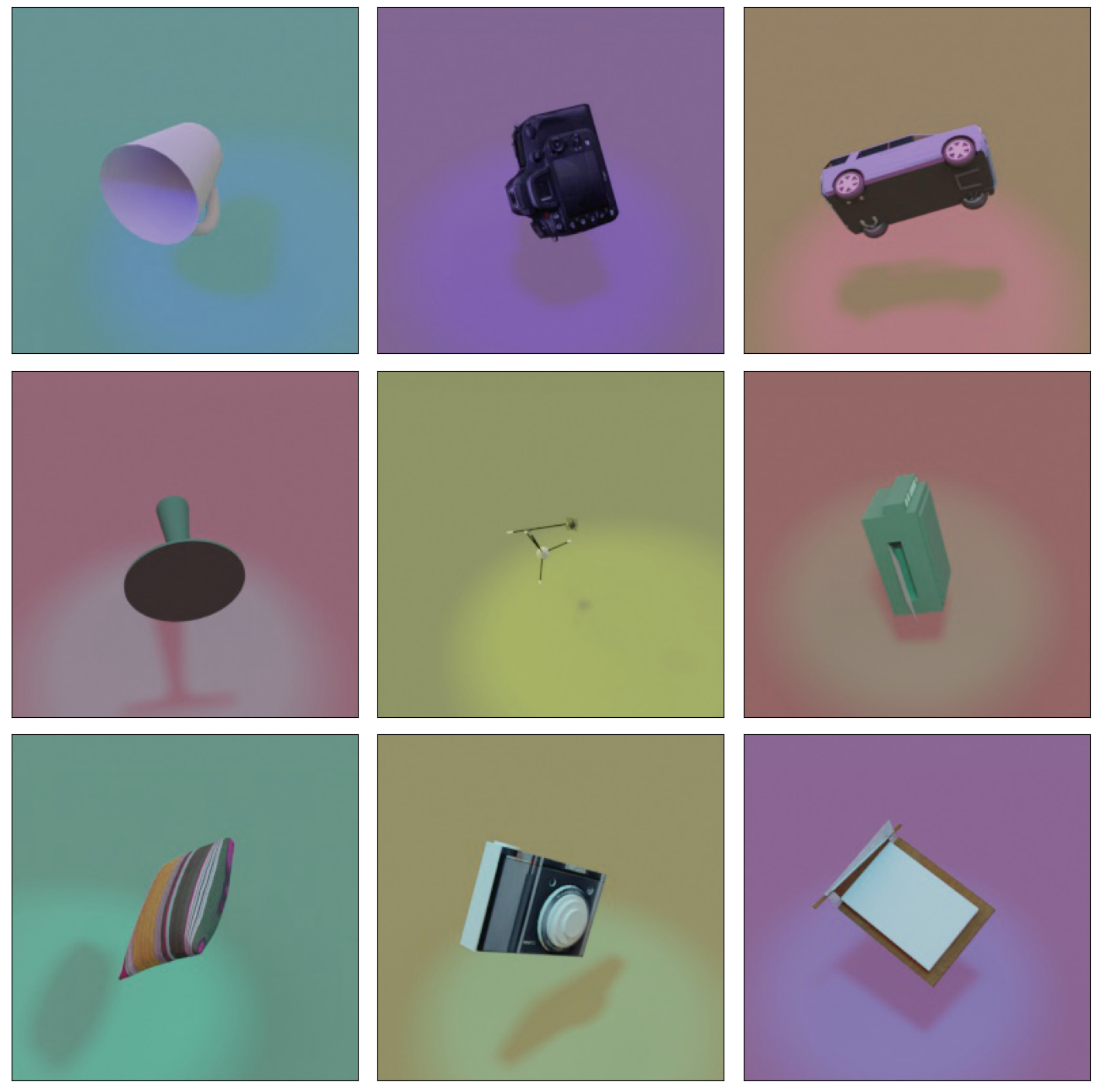}
    \caption{Image samples from 3DIEBench.}
    \label{fig:data_sample}
    \vspace{-0.2cm}
\end{figure}

\section{Creating a general predictor}

\begin{figure*}[!t]
    \centering
    \includegraphics[width=0.55\textwidth]{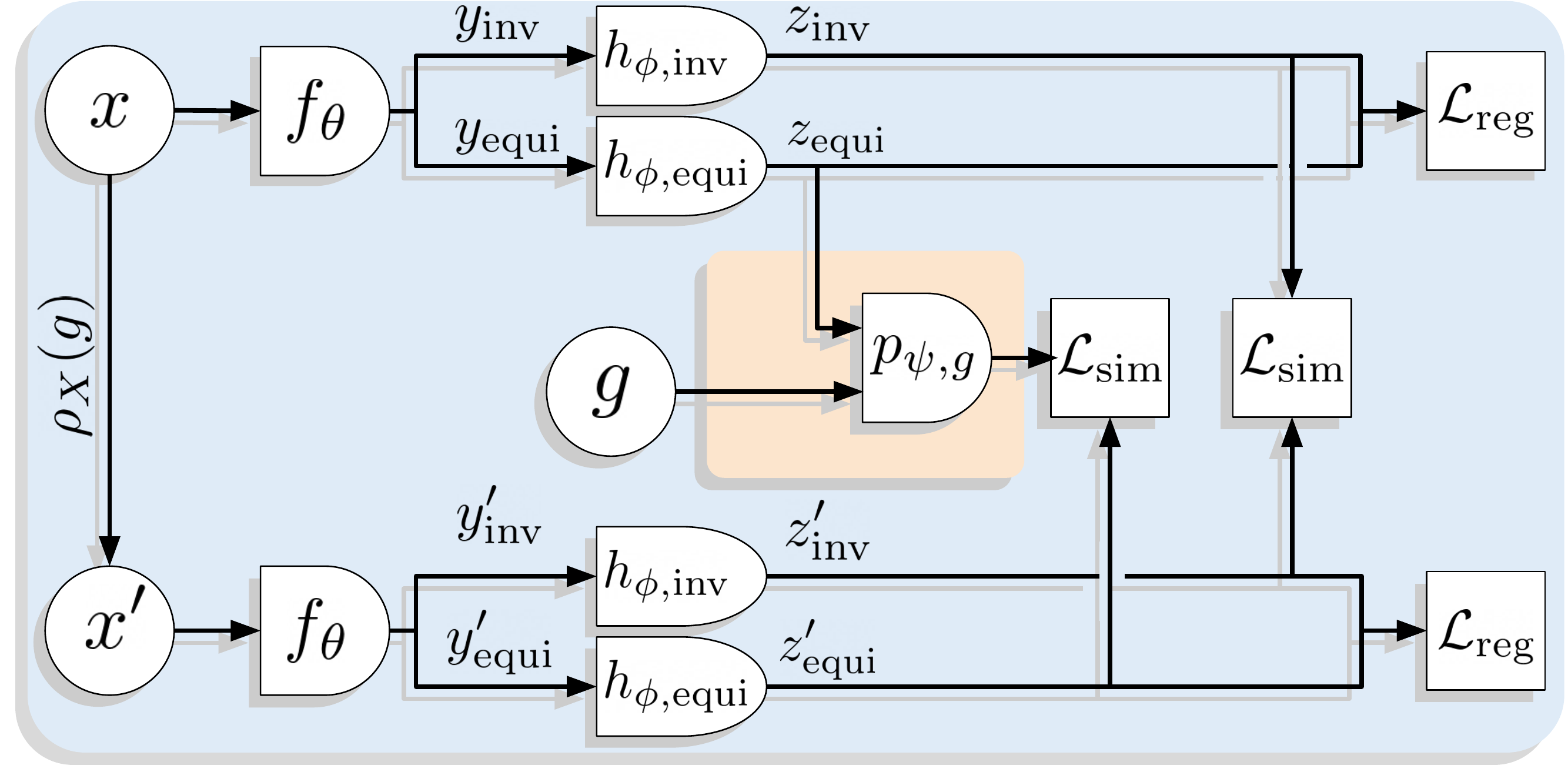}
    \hfill
    \includegraphics[width=0.44\textwidth,trim = 0cm -3.5cm 0cm 0cm]{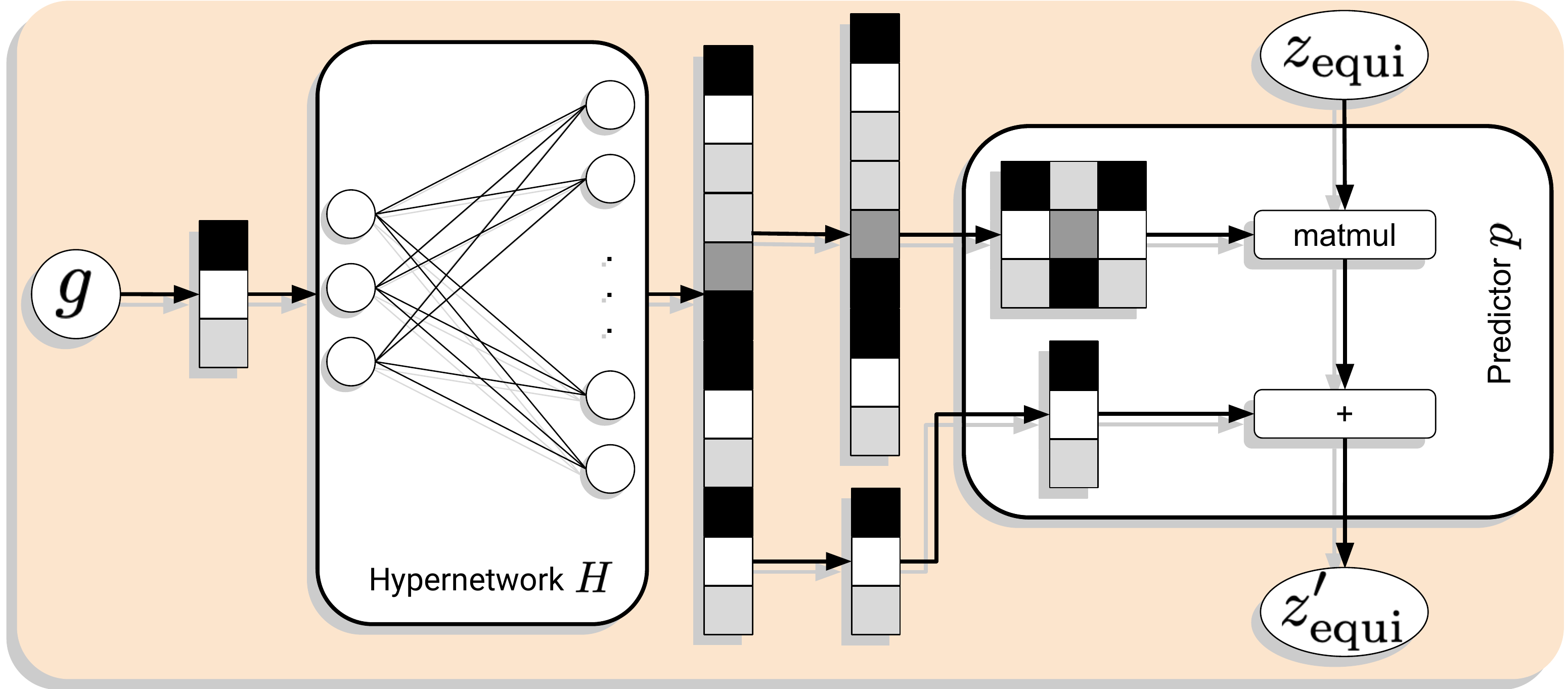}
    \caption{\textbf{Left} Schematic overview of the proposed architecture. Representations are split in an invariant and equivariant part, where the equivariant part is then fed through a predictor. \textbf{Right} Our hypernetwork-based predictor takes a group element as input and produces the weights associated to a transformation in latent space, alowing us to learn equivariant representations. }
    \label{fig:method}
\end{figure*}

\subsection{Background and notation}

\paragraph{Group actions}

A group consists of a set $G$ with a binary operation $\cdot : G\times G \rightarrow G$, which is associative, where there exists an identity element $e\in G$ such that $\forall g \in G,\; g\cdot e = g$ and $ e\cdot g = g$, and where every element $g \in G$ has an inverse $g^{-1}$ such that $g\cdot g^{-1} = e$ and $g^{-1}\cdot g = e$. 
Our focus is on 3D rotations, i.e. $SO(3)$, but we use quaternions to represent them, i.e. $Sp(1)$, for their ease of use. Hue changes for the floor and lighting are represented by the group $(\mathbb{R},+)$.

A group action of $G$ on a set $S$ is defined\footnote{This defines a left group action and a right group action can be defined analogously as $\alpha : S\times G \rightarrow S$} as a function $\alpha : G\times S \rightarrow S$ such that $\alpha(e,s) = s $ and which is compatible with the composition of group elements $\alpha(g,\alpha(h,x)) = \alpha(gh,x)$. If $\alpha$ is linear and acts on a vector space $V$ such as $\mathbb{R}^n$, it is called a group representation.
We then define a group representation as the map $\rho: G \rightarrow GL(V)$ such that $\rho(g) = \alpha(g,\cdot)$.
In practice, group representations describe how transformations are applied on our input data, as well as in our latent space. Considering an input image $x$, an augmented view $x'$ can be defined as $x' = \rho(g)\cdot x$, where $g$ describes the augmentation parameters.

\paragraph{Invariant self-supervised learning}
Starting from a dataset $\mathcal{D}$ with datum $d \in \mathbb{R}^{c\times h\times w}$, we generate two augmented views $x$ and $x'$ using any data augmentation strategy, as long as it preserves semantically meaningful information for a target downstream task. Both views are then fed through an encoder $f_\theta$ to obtain representations $y = f_\theta(x)$ and $y' = f_\theta(x')$. These representations are then fed through a projection head $h_\phi$ to obtain embeddings $z = h_\phi(y)$ and $z' = h_\phi(y')$. The goal is then to make the pairs of embeddings $(z,z')$ identical to learn embeddings that are invariant to the applied augmentations and extract meaningful information from the original images. When considering batches, we denote matrices by capital letters. For example, the matrix of embeddings will be $Z$, and $Z_i$ the $i$-th embedding.
While the loss is applied on the embeddings, representations are used in practice as it has been shown to increase performance~\citep{chen2020simple} and avoid complete invariance in the representations which can be detrimental~\cite{bordes2022guillotine}.
To make a link to the previously discussed group actions, we can consider the data augmentation strategy as a group representation $\rho_X$. We can then define $x$ and $x'$ as $x = \rho_X(g_1)\cdot d$, $x' = \rho_X(g_2)\cdot d$ and especially $x' = \rho_X(g)\cdot x$ with $g = g_1^{-1}\cdot g_2$. The goal of invariant self-supervised approaches is then to have $\forall x \in \mathcal{D}\; \forall g \in G,\; h_\phi(f_\theta(x)) = h_\phi(f_\theta(\rho_X(g)\cdot x))$.

\paragraph{Equivariant representations}
Given a group $G$ with representations $\rho_X$ and $\rho_Y$, we say that a function $f: X \rightarrow Y$ is equivariant with respect to $G$ if $\forall x \in X, \forall g \in G$ we have
\begin{equation*}
    f(\rho_X(g)\cdot x) = \rho_Y(g)\cdot f(x).
\end{equation*}
This means that a function $f$ is equivariant if it commutes with group transformations. It is worth noting that we can see invariance as a special case where $\rho_Y(g) = Id$. As we show in our experiments, this is a common failure mode of existing equivariant approaches.

While prior works have focused on forcing equivariance by the architecture of $f$~\citep{cohen2016group,cohen2018spherical}, we focus on a setting where this is not possible, and where we do not even know $\rho_X$. Indeed, in our constructed dataset, it is impossible to apply the transformation on 
the renderings, even though this was possible in the original 3D space. However, we still know the group elements that parametrized our transformation and are able to make use of them.
The goal is then to learn both $f$ and $\rho_Y$ in order to obtain representations that are as equivariant as possible to the original transformation.

\subsection{Our Method: SIE}

\paragraph{General architecture}

While we are placed in the joint-embedding framework described previously, we introduce a split in two of the representations before the projection head $h_{\phi}$. We separate $y$ (resp. $y'$) as $y_{\text{inv}}$ which contains information that is preserved by the transformation, i.e. invariant information, and $y_{\text{equi}}$ which contain information that was changed by the transformation, i.e., equivariant information. To illustrate, if $y$ is 512-dimensional, we use the first 256 dimensions for $y_{\text{inv}}$ and the 256 last for $y_{\text{equi}}$. We thus call our approach Split Invariant Equivariant (SIE).
Both parts are then fed through separate projection heads $h_{\phi,\text{inv}}$ and $h_{\phi,\text{equi}}$ to ensure that no information is exchanged between the two after the split. This gives us embeddings $z_{\text{inv}}$ and $z_{\text{equi}}$.
While we want the invariant embeddings $z_{\text{inv}}$ and $z'_{\text{inv}}$ to be identical, the equivariant embeddings $z_{\text{equi}}$ and $z'_{\text{equi}}$ should only be identical after the predictor $p_{\psi,g}$, which is parametrized by the transformation between the two views $g$. As such, $p_{\psi,g}$ is our learnable $\rho_Y(g)$ described previously. The representation split can also be interpreted as a single predictor on the whole representations where we force it to be the identity for certain dimensions of the representations ($z_{\text{inv}}$) and allow more flexibility on the rest of the dimensions ($z_{\text{equi}}$). This whole process is illustrated in figure~\ref{fig:method}.

We now discuss the loss function used to train SIE. We use {\mbox VICReg~\citep{bardes2021vicreg}} as our basis since it lends itself well to split representation.
In order to adapt its invariance criterion, we define our similarity criterion $\mathcal{L}_{\text{sim}}$ as 
\begin{equation*}
    \mathcal{L}_{\text{sim}}(u,v) = \| u - v \|_2^2 ,
\end{equation*}
which we use to match both our invariant and equivariant embedding pairs. In order to avoid a collapse of the representations, we use the original variance and covariance criterion to define our regularisation criterion $ \mathcal{L}_{\text{reg}}$ as 
\begin{align*}
    \mathcal{L}_{\text{reg}}(Z) &= \lambda_C\; C(Z) + \lambda_V\; V(Z), \quad\text{with}\\
     C(Z) &= \frac{1}{d}\sum_{i\neq j} Cov(Z)_{i,j}^2  \quad\text{and}\\
     V(Z) &=\frac{1}{d} \sum_{j = 1}^d \max\left( 0, 1- \sqrt{Var(Z_{\cdot,j})}\right) .
\end{align*}
The goal of the variance criterion $V$ is to ensure that all dimensions are used in the embeddings and the goal of the covariance criterion $C$ is to decorrelate the dimensions to spread out the information in the embeddings.
We are now ready to introduce our final criterion as
\begin{align*}
    \mathcal{L_{\text{SIE}}} (Z,Z') =&   \mathcal{L}_{\text{reg}}(Z') + \mathcal{L}_{\text{reg}}(Z) + \textcolor{gray}{\lambda_V V(p_{\psi,g_i} (Z_{i,\text{equi}}))+}\\
    & \lambda_{\text{inv}} \frac{1}{N} \sum_{i=1}^N   \mathcal{L}_{\text{sim}}(Z_{i,\text{inv}},Z'_{i,\text{inv}}) +\\ 
   & \lambda_{\text{equi}}  \frac{1}{N} \sum_{i=1}^N \mathcal{L}_{\text{sim}}(p_{\psi,g_i}(Z_{i,\text{equi}}),Z'_{i,\text{equi}}).
\end{align*}

Notice that the regularisation criterion $\mathcal{L}_{\text{reg}}$ is applied to the whole embeddings and not separately for the invariant and equivariant part. While this does not matter for the variance criterion, it allows the covariance criterion to decorrelate information between the invariant and equivariant parts which is consistent with our goal.
We also add another variance criterion (in \textcolor{gray}{gray}) on the output of the predictor to help stabilize training. Its goal is to avoid a fully collapsed predictor since it normalizes the predictions. While this helps to stabilize the beginning of the training it does not impact final performance. However, without it, some runs never learn useful representations and fall back to VICReg's behaviour. It is thus an optional yet recommended component.
We use $\lambda_{\text{inv}} = \lambda_V = 10$,$\lambda_{\text{equi}} = 4.5$, and $\lambda_C = 1$ in our experiments.

\paragraph{Predictor architecture}

Previous works have employed predictors in self-supervised learning, for different purposes. In~\cite{grill2020byol,chen2020simsiam} the predictor used is purely deterministic and does not depend on the target. As such its only solution is to converge to the identity, giving it a limited role in introducing any level of equivariance. 
Recent works have introduced predictors that depend on the transformation between the two views of the input and have converged to a linear transformation in general~\citep{devillers2022equimod}, or only for 3D rotations~\citep{park_learning_2022}. 
As we study in supplementary section~\ref{sec:pred-collapse}, this kind of architectures can ignore the transformation parameters and collapse back to behaviours associated with invariance based methods, i.e. $p_{\psi,g} = Id$.

To ensure that the transformation parameters are taken into account by the predictor, we introduce a predictor architecture based on hypernetworks~\citep{ha2016hypernetworks}. The idea is to use a neural network $H : G \rightarrow \mathbb{R}^w$ which takes as inputs our transformation parameters $g$ and outputs the weights parametrizing the predictor. As an example, if our representations are $d$-dimensional and our desired predictor is a linear transformation, $H$ outputs a $d^2$-dimensional weight vector that can be reshaped and used for our prediction. This process is described in figure~\ref{fig:method}.
This architecture gives us almost complete freedom into the predictor itself, but for simplicity we consider a linear transformation. Similarly, to ensure that $g$ is not ignored, $H$ is a linear transformation. It is very important to not have a bias parameter in $H$ otherwise it could collapse to a predictor akin to SimSiam or BYOL's predictors by setting all weights relatd to $g$ to 0.
To summarize, $p_{\psi,g}$ is a linear transformation with weights $\psi$ defined as the output of $H$, i.e. $ \psi = H(g)$. We thus have $p_{\psi,g}\left(z_{\text{equi}}\right) = \texttt{reshape}\left(H(g), d \times d \right) z_{\text{equi}} $, where $\texttt{reshape}\left(v, d \times d \right)$ reshapes a vector $v$ into a { $d\times d$ matrix}.

\begin{table*}[!t]
    \centering
    \caption{Quantitative evaluation of learned representations on invariant (classification) and equivariant (rotation prediction, color prediction) tasks. Equivariant methods are trained to be equivariant to rotation, but no constraint is given for color. For each family of methods we highlight the best value in bold. We see that while SIE suffers from a small drop in classification performance, it outperforms all equivariant methods even when using the same predictor.$^ \dagger$ We train a supervised baseline for each evaluation.}
     \begin{tabular}{lccccccccc}
        \toprule
        Evaluation & \multicolumn{3}{c}{Classification (top-1)} & \multicolumn{3}{c}{Rotation prediction ($R^2$)} & \multicolumn{3}{c}{Color prediction ($R^2$)}   \\
        \cmidrule(lr){2-4} \cmidrule(lr){5-7}\cmidrule(lr){8-10}
         Representation part (if applicable) & All & Inv. & Equi.  & All & Inv. & Equi. & All & Inv. & Equi. \\ 
         \midrule
         Supervised$^ \dagger$ & 87.47 & & & 0.76 & & & 0.99  \\
         \midrule
         \multicolumn{2}{l}{\textit{\textcolor{gray}{Invariant and parameter prediction methods}}}\\
         VICReg & 84.74 & & &0.41 &  & & 0.06 \\
         VICReg, $g$ kept identical & 72.81  & & &0.56 &  & & 0.25  \\
         SimCLR & \textbf{86.73} & & & 0.50 & & & 0.30  \\
         SimCLR, $g$ kept identical & 71.21  & & & 0.54 & & & 0.83  \\
         SimCLR + AugSelf & 85.11 & & & \textbf{0.75} & & & 0.12   \\
         \midrule
         \textit{\textcolor{gray}{Equivariant methods}}\\
         Only Equivariance (Original predictor) & 86.93  & & & 0.51 & & & 0.23   \\
         Only Equivariance (Our predictor) & 86.10 & & & 0.60 & & & 0.24 \\
         EquiMod (Original predictor) & \textbf{87.19} & & & 0.47 & & & 0.21  \\
         EquiMod (Our predictor) & \textbf{87.19} & & & 0.60 & & & 0.13  \\
         SIE (Ours)  & 82.94 & 82.08  & 80.32  & \textbf{0.73} & 0.23 & 0.73 & 0.07 & 0.05 & 0.02  \\
         \bottomrule
    \end{tabular}
    \label{tab:quantitative}
\end{table*}
\section{Experiments}

\subsection{Methods and protocols}

\paragraph{Compared methods} We compare our approach to VICReg~\cite{bardes2021vicreg} and SimCLR~\cite{chen2020simple} to have baselines for invariant self-supervised methods. We consider both the scenarios where they have to be invariant about $g$ as well as a scenario where $g=0$ and where we apply standard image augmentations, following the protocol of~\cite{grill2020byol}. The goal is to see if we learn some information about the object pose by considering different poses as different samples instead of augmented views.
We also compare our approach to SimCLR+AugSelf~\cite{lee_improving_2021} as a parameter prediction method. It is trained to predict $g$, but since it does not provide a transformation in embedding space ($\rho_{Y}(g)$) it cannot be considered equivariant and is mostly included for completeness.
Finally we compare ourselves to SEN~\cite{park_learning_2022} and EquiMod~\cite{devillers2022equimod} for equivariant methods. We consider them both with their original predictor as well as with our hypernetwork based predictor to demonstrate both its benefits as well as the benefits of the invariant-equivariant split.
For SEN, we use the same contrastive loss as SimCLR instead of the original triplet loss to limit hyperparameter tuning. For clarity we label this change as Only Equivariance.

\paragraph{Training protocols}

All methods use a ResNet-18~\citep{he2016resnet} as their encoder and a three layer MLP as projection head. To obtain asymptotic behaviours they are all trained for 2000 epochs using the Adam optimizer~\citep{kingma2014adam}, with learning rate $10^{-3}$ and default $\beta$ parameters. We give more details on the pretraining strategies in supplementary section~\ref{sec:protocol}.

\begin{figure*}[!th]
    \centering
    \includegraphics[width=\textwidth]{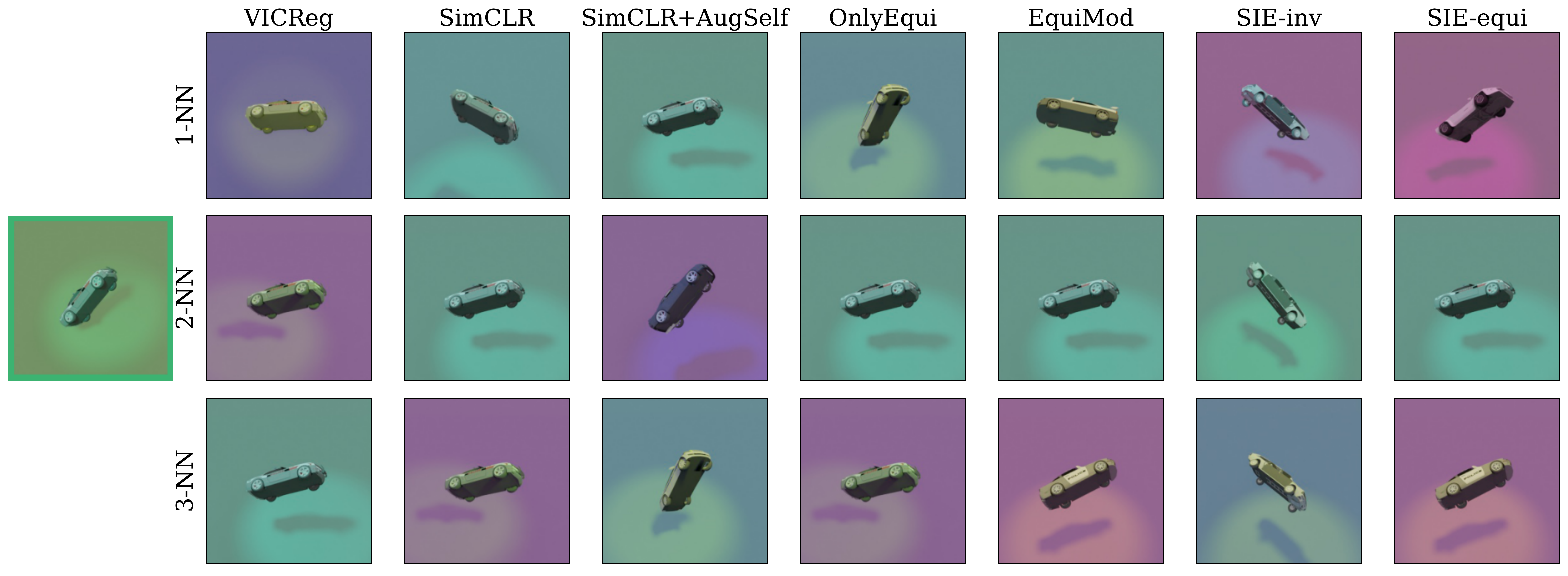}
    \vspace{-0.5cm}
    \caption{Retrieval of nearest representations. Starting from the representation associate to the object in the \textcolor{green}{green} frame on the left, we compute its nearest neighbours for all considered methods and show the 3 closest corresponding images.}
    \label{fig:sample_reprs}
    \vspace{-0.2cm}
\end{figure*}

\subsection{Representations evaluation}

\paragraph{Metrics and protocol} As is common practice in self-supervised learning we start by evaluating the quality of the representations on dowsntream tasks. We use linear classification on top of frozen representations as our representative invariant task. It is worth noting that this is not purely invariant since some information about the transformation is helpful in practice~\cite{bordes2022guillotine}. For our representative equivariant task, we use rotation prediction. The representations from two transformed views of the same object are fed through a 3 layer MLP that is trained to regress the rotation between the two. We also include linear regression of the floor and spot hue to study any side effects on a task that is comparatively simple. We give more details on the evaluation protocols in supplementary section~\ref{sec:protocol}.

\paragraph{Quantitative results} Our results are summarized in table~\ref{tab:quantitative}. We first notice that invariant self-supervised learning methods offer classification accuracies that are very close to the supervised baselines, but they offer worse performance in rotation prediction. When considering rotated objects as different instances and not as a transformation, we notice an increase of performance on rotation prediction at the cost of classification performance. Overall, VICReg offers a higher level of invariance than SimCLR across all metrics.
AugSelf yields a significant boost over the SimCLR baseline for rotation prediction performance, albeit with a small drop in classification. Its pretraining task is identical to our evaluation which explains why its performance is close to the supervised baseline.
Looking at equivariant methods, we see that both EquiMod and Only Equivariance perform very well on classification but offer no increase in performance for rotation prediction compared to SimCLR which serves as their base. This would suggest that their original predictor does not induce more equivariance that the implicit equivariance offered by the projection head. We discuss this behaviour below.

When using our hypernetwork based predictor with EquiMod or Only Equivariance, we notice a clear boost in performance in rotation prediction, showing that it is able to improve the performance on equivariance related tasks. The performance is however still far from the supervised baseline. When combining both this predictor architecture and our split representations, SIE is able to further improve performance on rotation prediction, but by incurring a small performance drop compared to its VICReg base.
Looking into more details about SIE's results, we see that the equivariant part of the representations still contains a significant amount of information that is helpful for classification. The rotations that we apply can be so extreme that knowing the nature of the object is important to learn how to apply a rotation. When looking at the invariant part of the representations, we see that it achieves the lowest performance, showing that SIE learned to most invariant representations. As for color, we also obtain almost perfect invariance, highlighting again the invariance of our representations to transformations that have no incentive to be preserved.
Overall, both our predictor architecture and invariant-equivariant split help to greatly improve performance on equivariance related tasks, while also obtaining the highest level of invariance when desired. We study the task of learning representations that are both equivariant to rotation and color in supplementary section~\ref{sec:rot-col}, where we are able to show increased performance in color prediction with only slight drops in performance for rotation prediction.

\paragraph{Implicit equivariance of the projection head} While a significant part of the performance on equivariance related tasks can be attributed to the predictor, even invariant methods have various level of performance. This can be explained by the use of the projection head which absorbs some of the bias from the invariance criterion~\citep{bordes2022guillotine,chen2020simple}. However, this is not enough to achieve satisfactory performance, nor does it give a way to steer the latent space. Nonetheless, even predictor based methods can benefit from it and it remains an easy way to improve quantitative performance. We provide a more in depth analysis of the role of the projection head in supplementary section~\ref{sec:emb-perf}, where we see that SIE suffers from the smallest drops in performance after the projection head, matching the performance on representations of EquiMod, whereas EquiMod falls to a level of performance similar to VICReg before its projection head.

\paragraph{Predictor collapse to the identity}
The predictor architecture that was originally used in~\cite{park_learning_2022,devillers2022equimod} was a linear layer which takes as input the concatenation of the representations and the transformation parameters. 
This means that it can simply choose to ignore the transformation parameters by setting the appropriate weights to 0.
This behaviour is also accompanied by a collapse to the identity of the predictor, since solving the invariant task is easier than learning equivariant representations.
This happens in practice for every method, and is the reason why the performance with this predictor architecture is similar to what an invariant method gives. Confer supplementary section~\ref{sec:pred-collapse} for a study of this phenomenon.

\paragraph{Qualitative results} In order to visualize the information present in the representations, we perform a retrieval of nearest representations on our validation set. We expect the representations of invariant methods to contain similar objects in various poses but representations of equivariant methods to contain objects in similar poses to the queried representations. As we can see in figure~\ref{fig:sample_reprs} all methods lead to nearest-neighbours in similar poses as the queried object except for the invariant part of the representations from SIE. Nonetheless, for SimCLR, Only Equivariance and EquiMod, the nearest neighbour is not in a similar pose as the queried object, highlighting an imperfectly learned equivariant mapping.
Invariant methods preserve transformation related information in a way that is stronger than expected due to the implicit equivariance introduced by the projection head. 
We reproduce the same figure on embeddings in supplementary section~\ref{sec:emb-perf} and notice that invariant methods do not lead to nearest neighbours in similar poses, whereas equivariant methods tend to perform better, especially SIE. 
We do notice that all objects are cars, which would suggest that the class information is still present, confirming our quantitative results.

\begin{table*}[!t]
    \centering
    \caption{Quantitative evaluation of the predictor, using PRE, MRR and H@k. We specify the source dataset on which embeddings are computed (train or val), and when necessary the dataset used for retrieval (train,val or all). We see that on all metrics SIE outperforms by a large margin EquiMod and Only Equivariance with our hypernetwork-based predictor.}
    \begin{tabular}{lccccccccc}
        \toprule
        Method & \multicolumn{3}{c}{PRE ($\downarrow$)} & \multicolumn{2}{c}{MRR ($\uparrow$)} & \multicolumn{2}{c}{H@1 ($\uparrow$)} & \multicolumn{2}{c}{H@5 ($\uparrow$)}\\
        \cmidrule(lr){2-4} \cmidrule(lr){5-6} \cmidrule(lr){7-8}\cmidrule(lr){9-10}

        & train-train & val-val & val-all & train & val & train & val & train & val   \\
        \midrule
          EquiMod & 0.47 & 0.48 & 0.48 & 0.17 & 0.16 & 0.06  & 0.05 & 0.24 & 0.22  \\
          Only Equivariance  &  0.47 & 0.48 & 0.48 & 0.17 & 0.17 & 0.06  & 0.05 & 0.24 & 0.22   \\
          SIE (Ours)  & \textbf{0.26} &  \textbf{0.29} & \textbf{0.27} & \textbf{0.51} & \textbf{0.41}  & \textbf{0.41} & \textbf{0.30}  & \textbf{0.60 }& \textbf{0.51 } \\
         \bottomrule
    \end{tabular}
    \label{tab:pred-eval}
\end{table*}

\subsection{Predictor evaluation}

\paragraph{Metrics} In order to evaluate the quality of the predictors of equivariant methods, we adapt commonly used metrics~\cite{kipf2019contrastive,park_learning_2022} such as the Mean Reciprocal Rank (MRR) and Hit Rate at k (H@k)  to our multi object setting. Starting from source and target poses for an object, we feed the source embeddings through the predictor and look at the nearest neighbours of the predicted embedding, i.e, $k\text{-NN}(p_{\psi,g}(Z_{\text{source,equi}}))$. The MRR is then defined as the average reciprocal rank of the target embedding in this nearest neighbour graph. H@k is a harder measure which returns 1 if the target embedding is in the $k$-NN graph around the predicted embedding and 0 otherwise. To adapt them to our setting, we only look for nearest neighbours among the views of the same object. This helps to avoid penalizing the predictor when the retrieval yields an incorrect object that is still in a pose close to the correct one. 
To get a finer understanding of the predictor quality we introduce Prediction Retrieval Error (PRE). It is computed by taking the nearest neighbour of a predicted embedding and computing the distance between its rotation $q_1 \in \mathbb{H}$ and the target rotation $q_2$ as $d = 1 - <q_1,q_2>^2$. Averaged over the whole dataset, this gives us a measure of the prediction's quality. Confer supplementary section~\ref{sec:protocol} for details.

\begin{figure}[!t]
    \centering
    \includegraphics[width=\columnwidth]{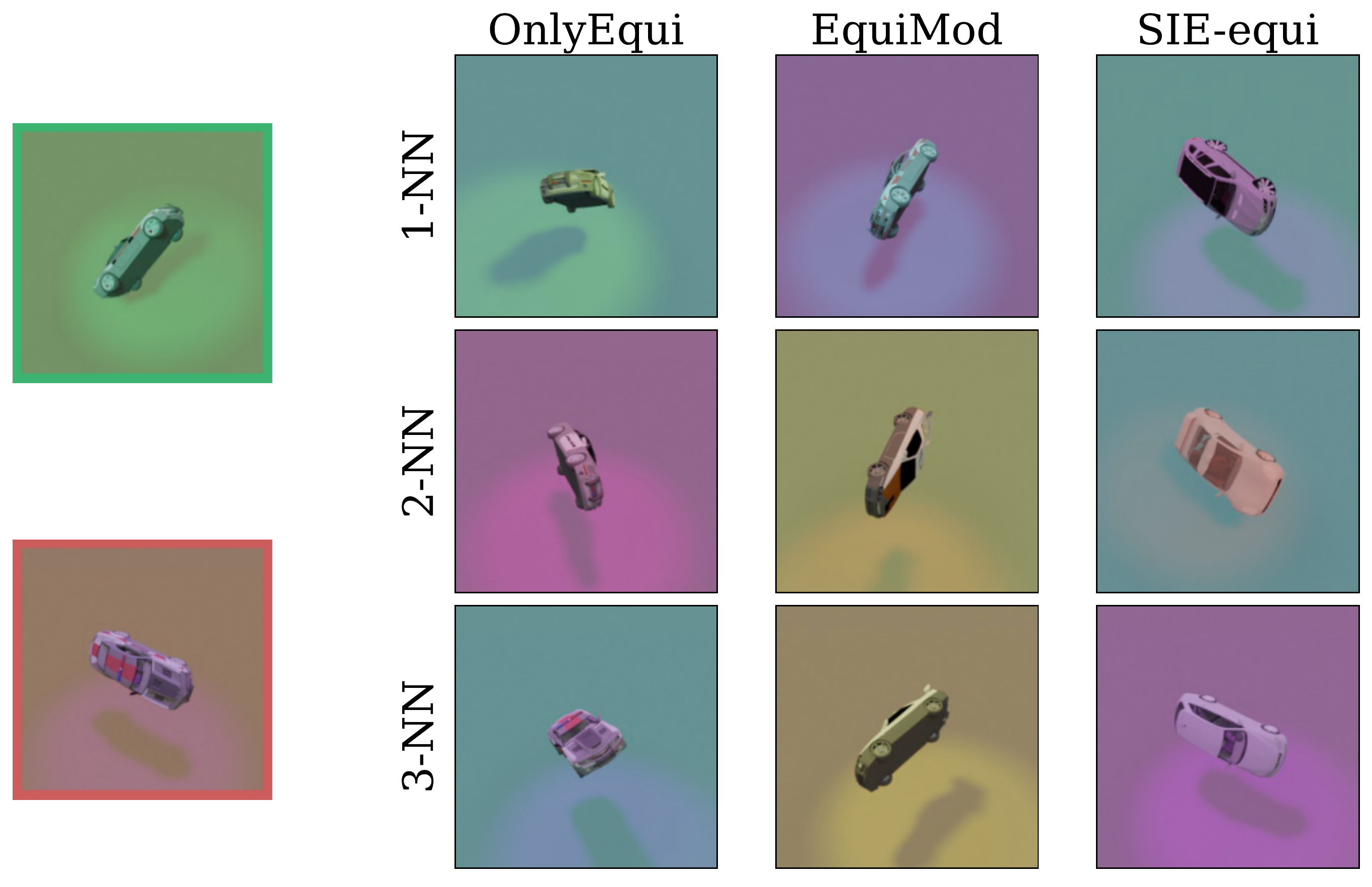}
    \caption{Retrieval of predicted embeddings. Starting from the embedding of the object in the \textcolor{green}{green} frame and having as target the embedding of object in the \textcolor{red}{red} frame, we look at the nearest neighbours of the predicted embddings. SIE leads to predicted embeddings of objects in the same pose as the target, contrasting with other methods.}
    \label{fig:sample_pred}
    \vspace{-0cm}
\end{figure}

\paragraph{Quantitative results} As we can see in table~\ref{tab:pred-eval}, no matter the considered dataset, SIE consistently outperforms EquiMod and Only Equivariance, achieving $0.29$ PRE on the validation set compared to $0.48$ for EquiMod and Only Equivariance. The results are similar with MRR and H@1/5 where on both the training and validation set SIE outperforms EquiMod and Only Equivariance. To interpret better what this means, the H@1 of $0.3$ for SIE means that $30\%$ of the time the nearest neighbour is the target embedding, whereas this is only true $5\%$ of the time for EquiMod and Only Equivariance. This is only slightly better than random which would be $2\%$. Since the same predictor is used for all methods, this highlights the importance of using split representations.

\paragraph{Qualitative results} To give a clearer picture of the predictor's influence on embedding, we show the closest neighbours of predicted embedding in figure~\ref{fig:sample_pred}. Starting from a pair of embeddings, we apply the predictor on the starting embedding with the goal of rotating it to the pose of the target embedding. When retrieving the nearest neighbours of the predicted embeddings, we expect them to be objects in a similar pose as the target. For both Only Equivariance and EquiMod, we see that the predicted embeddings are dissimilar to the target embedding, whereas for SIE, we do find other cars in the same pose.

While both quantitative and qualitative results would suggest that the predictor is of lower quality in EquiMod and Only Equivariance compared to SIE, these results must be interpreted with care. Since our visualization and metrics rely on nearest neighbour retrieval on the whole dataset, if equivariant information contributes less to the norm of the embeddings compared to invariant information then it may not be captured well. Nonetheless, this still shows that the invariant-equivariant split in SIE plays a significant role in making equivariant information easily accessible.
As mentioned previously, the drop in quality when going from the representations to the embeddings can be partially attributed to the implicit equivariance induced by the projection head.

\section{Limitations}

While we have shown improved performance over existing methods both thanks to our hypernetwork-based predictor and split representations, SIE currently requires knowledge about the group elements. This can limit its applicability in settings where they are unknown or where only partial information is available. While existing works also suffer from this limitations, removing the need for this knowledge is an important future line of work.\\
We have also shown that using split representations significantly helps equivariance-related performance, however this comes at a small cost in invariance performance. When optimizing for different tasks a trade-off has to be made for performance and different methods will be optimal for a given use-case. SIE gives a satisfactory trade-off, maximizing equivariance performance while preserving most of the invariance performance, but other choices may be better suited for different targeted downstream tasks.

\section{Conclusion}

We have introduced a method to learn both invariant and equivariant representations based on self-supervised learning. By introducing 3DIEBench, we create an experimental setting which is more challenging than existing datasets to learn equivariant representations and that also enables us to evaluate on image classification, a task that can be assimilated to invariant representations.
By using a predictor based on a  linear hypernetwork and by splitting representations in an invariant and equivariant part, SIE is able to beat existing equivariant methods on both qualitative and quantatitative metrics. Reproducing previous works with our hypernetwork-based predictor further enabled us to show the positive impact on performance of both the predictor design and the use of split representations.
We hope that SIE can serve as a basis towards the design of equivariant methods in more complex settings, enabling the use of self-supervised learning to learn richer representations.

\section{Acknowledgments}
We wish to thank Adrien Bardes for insightful discussions as well as inspiring the first experiments that led this work.
We also want to thank Randall Balestrierio,  Gr\'{e}goire Mialon, Pascal Vincent, Florian Bordes, Nicolas Ballas, Mido Assran and Yubei Chen for insightful discussions throughout the development of this work.

\bibliography{biblio}
\bibliographystyle{icml2023}

\newpage
\appendix
\onecolumn
\renewcommand{\thefigure}{S\arabic{figure}}
  \renewcommand{\thetable}{S\arabic{table}}
\setcounter{figure}{0}
\setcounter{table}{0}

\section{Exact training and evaluation protocols\label{sec:protocol}}
\subsection{Pretraining}

As previously mentioned, all methods are trained for 2000 epochs using a resnet-18 encoder and MLP projection head. We use a batch size of 1024 with the Adam optimizer with learning rate $10^{-3}$, $\beta_1 = 0.9$ , $\beta_2 = 0.999$. All experiments are done using 4 NVIDIA V100 GPUs and take around 24 hours. As we discuss in supplementary section~\ref{sec:ablations}, shorter training regimens can be used and lead to similar performance. We discuss method-specific hyperparameters.

\paragraph{Supervised} In the supervised baselines we used the same protocol and used the same evaluation heads that are used when evaluating self-supervised approaches on top of a resnet-18, see next subsection for details. We train for 2000 epochs to also obtain asymptotic results and to provide better upper bounds on performance.

\paragraph{VICReg} We use a projection head with intermediate dimensions 2048-2048-2048, as well as loss weights $\lambda_{inv} = \lambda_{var} = 10$ and $\lambda_{cov} = 1$.

\paragraph{SimCLR, Only Equivariance} We use a projection head with intermediate dimensions 2048-2048-2048, and temperature $\tau=0.1$ for the loss.

\paragraph{SimCLR + AugSelf} We use a projection head with intermediate dimensions 2048-2048-2048, and temperature $\tau=0.1$. For the parameter prediction head, we use a MLP with intermediate dimensions 1024-1024-4. We weigh the two losses (SimCLR and parameter prediction) equally.

\paragraph{EquiMod} We follow the original protocol and use projection heads with intermediate dimensions 1024-1024-128. We use these dimensions to coincide with the ones used for SIE, giving us a fair comparison. We use $\tau=0.1$ as our loss temperature, and weigh the two losses (invariance and equivariance) equally. We tried with different weights and found the original equivariance weight of 1 to work best.

\paragraph{SIE} For both our invariance and equivariant projection heads we use intermediate dimensions 1024-1024-1024. We use $\lambda_{\text{inv}} = \lambda_V = 10$, $\lambda_{\text{equi}} = 4.5$, and $\lambda_C = 1$ in our experiments. See supplementary section~\ref{sec:ablations} for ablations on these parameters.

\subsection{Evaluation}

\paragraph{Classification} Following common protocols we train a linear classification head on top of frozen representations for 300 epochs using a batch size of 256. We rely on the Adam optimizer with learning rate $10^{-3}$, $\beta_1 = 0.9$ , $\beta_2 = 0.999$. We use a cross entropy loss to train our classifier. Performance is then reported on the validation set. 

\paragraph{Rotation prediction} We train a MLP with intermediate dimensions 1024-1024-4 on top of our frozen encoder. Its inputs are pairs of representations that are concatenated. We train for 300 epochs using a batch size of 256. We rely on the Adam optimizer with learning rate $10^{-3}$, $\beta_1 = 0.9$ , $\beta_2 = 0.999$. We use a MSE loss as our regression loss.Performance is then reported on the validation set using $R^2$, which contains unseen objects with the same pose distribution as seen during training.

\paragraph{Color prediction} We train a linear regression head on top of frozen representations.Its inputs are pairs of representations that are concatenated. We train it for 50 epochs using a batch size of 256. We rely on the Adam optimizer with learning rate $10^{-3}$, $\beta_1 = 0.9$ , $\beta_2 = 0.999$. We use a MSE loss as our regression loss. Performance is then reported on the validation set using $R^2$.

\section{Ablations\label{sec:ablations}}

\begin{table}[!t]
    \centering
    \caption{Ablations on rotation prediction performance. \textbf{Top-left} Evaluation for different predictor architectures. \textbf{Top-right} Evaluation for different equivariance application methods. \textbf{Bottom} Influence of training duration on performance.}
    \hfill
    \begin{tabular}{lcc}
        \toprule
        Method &Rotation prediction ($R^2$) & Parameters   \\
        \midrule
          No-Invariance & 0.44 & 0   \\
          Linear  & 0.38 & 1M  \\
          MLP  & 0.38 & 4M \\
          Hypernetwork  & 0.73 & 4M  \\
         \bottomrule
    \end{tabular}
    \hfill
    \begin{tabular}{lc}
        \toprule
        Method &Rotation prediction ($R^2$)    \\
        \midrule
          Only Invariance & 0.41   \\
         Only Equivariance  & 0.60  \\
          VICReg-EquiMod  & 0.67  \\
          Split (two proj.)  & 0.73  \\
         \bottomrule
    \end{tabular}
    \hfill
    \vspace{0.1in}
    \begin{tabular}{lcccc}
        \toprule
        Method &500 ep. & 1000 ep.& 1500 ep. & 2000 ep.    \\
        \midrule
          VICReg & 0.44 & 0.43 & 0.42 & 0.41   \\
          SimCLR  & 0.46 & 0.48 & 0.50 & 0.50  \\
          Only Equivariance  & 0.57 & 0.61 & 0.59 & 0.60   \\
          EquiMod  & 0.57 & 0.60 & 0.61 & 0.60  \\
          SIE  & 0.68 & 0.72 & 0.73 &  0.73  \\
         \bottomrule
    \end{tabular}
    \label{tab:ablations-rot}
\end{table}
\begin{table}[!t]
    \centering
    \caption{Ablations on classification performance. \textbf{Top-left} Evaluation for different predictor architectures. \textbf{Top-right} Evaluation for different equivariance application methods. \textbf{Bottom} Influence of training duration on performance.}
    \hfill
    \begin{tabular}{lcc}
        \toprule
        Method & Top-1 accuracy (\%) & Parameters   \\
        \midrule
          No-Invariance & 69.04  & 0   \\
          Linear  & 82.24  & 1M  \\
          MLP  & 81.38 & 4M \\
          Hypernetwork  & 82.94 & 4M  \\
         \bottomrule
    \end{tabular}
    \hfill
    \begin{tabular}{lc}
        \toprule
        Method & Top-1 accuracy (\%)    \\
        \midrule
          Only Invariance &  84.74  \\
         Only Equivariance  & 80.98  \\
          VICReg-EquiMod & 84.10  \\
          Split (two proj.)  & 82.94  \\
         \bottomrule
    \end{tabular}
    \hfill
    \vspace{0.1in}
    \begin{tabular}{lcccc}
        \toprule
        Method &500 ep. & 1000 ep.& 1500 ep. & 2000 ep.    \\
        \midrule
          VICReg & 83.06 & 84.34 & 84.83 & 84.74    \\
          SimCLR  & 84.39  & 85.67 & 86.58 &  86.73  \\
          Only Equivariance  & 82.75 & 85.50 & 85.88 & 86.10    \\
          EquiMod  & 84.80 & 85.99 & 86.33 & 87.19   \\
          SIE  & 77.59 & 81.05 & 82.12 & 82.94    \\
         \bottomrule
    \end{tabular}
    \label{tab:ablations-classif}
\end{table}

\begin{table*}[!t]
    \centering
    \caption{Quantitative evaluation of the predictor, using PRE, MRR and H@k. We specify the source dataset on which embeddings are computed (train or val), and when necessary the dataset used for retrieval (train,val or all).}
    \begin{tabular}{lccccccccc}
        \toprule
        Method & \multicolumn{3}{c}{PRE ($\downarrow$)} & \multicolumn{2}{c}{MRR ($\uparrow$)} & \multicolumn{2}{c}{H@1 ($\uparrow$)} & \multicolumn{2}{c}{H@5 ($\uparrow$)}\\
        \cmidrule(lr){2-4} \cmidrule(lr){5-6} \cmidrule(lr){7-8}\cmidrule(lr){9-10}

        & train-train & val-val & val-all & train & val & train & val & train & val   \\
        \midrule
          EquiMod & 0.47 & 0.48 & 0.48 & 0.17 & 0.16 & 0.06  & 0.05 & 0.24 & 0.22  \\
          VICReg-EquiMod  &  0.36 & 0.37 & 0.37 & 0.36 & 0.29 & 0.25  & 0.18 & 0.46 & 0.39   \\
          SIE (Ours)  & \textbf{0.26} &  \textbf{0.29} & \textbf{0.27} & \textbf{0.51} & \textbf{0.41}  & \textbf{0.41} & \textbf{0.30}  & \textbf{0.60 }& \textbf{0.51 } \\
         \bottomrule
    \end{tabular}
    \label{tab:pred-eval-ablations}
\end{table*}

We study more carefully the impact of each component of SIE in tables~\ref{tab:ablations-rot},~\ref{tab:ablations-classif} and~\ref{tab:pred-eval-ablations}.

\paragraph{Rotation prediction} We can see that having either no invariance criterion on a part of the representation, or using a linear or MLP (4 layers) predictor leads to performance in the ballpark of VICReg. However, the use of the hypernetwork-based predictor significantly boosts performance, without necessarily increasing the parameter count of the model. We also see that the split representations lead to optimal representations, compared to using two projection heads on the full representations like EquiMod, or using only equivariance. For fairness, we used the same grid of equivariance weights for the VICReg-EquiMod scenario as we did for the split representations.

When looking at the performance for different training schedules, we see that performance plateaus after 1000 epochs, suggesting that shorter training regimens will lead to similar results for a much more manageable compute cost.

\paragraph{Classification} We see that the performance is in general not dependent on the predictor architecture. Even though we notice a small drop for the MLP predictor, the hypernetwork achieves similar performance as the linear predictor, which performed significantly worse on the rotation prediction task. Interestingly, using VICReg-EquiMod does not lead to a significant drop in performance compared to classical VICReg, whereas using only equivariance or split representations led to bigger drops.

Contrary to what we found for rotation prediction, training for longer always improve classification performance, even when going from 1500 to 2000 epochs. We also notice that SIE is the method that benefits the most from longer training, gaining over 5 points in top-1 accuracy going from 500 to 2000 epochs, compared to less than $2.5$ for EquiMod. This indicates that longer training are beneficial in general, even if they are not for rotation prediction.

\paragraph{Choice of base SSL criterion} As we have seen in table~\ref{tab:quantitative}, SimCLR's criterion leads to less invariant representations compared to VICReg's. However when looking at tables~\ref{tab:ablations-rot} and ~\ref{tab:ablations-classif} we see that when using VICReg's criterion, EquiMod seems to perform better on equivariant tasks compared to the base method using SimCLR's criterion. This shows an opposite behaviour for equivariant performance compared to the default versions of SimCLR and VICReg.
We also evaluate the predictor quality for VICReg-EquiMod in table~\ref{tab:pred-eval-ablations}. We can see that while VICReg-EquiMod achieves better performance than the classical EquiMod, SIE still outperforms it by a significant margin. This further demonstrates the usefulness of splitting representations.

\clearpage
\section{Performance on embeddings\label{sec:emb-perf}}

In order to better understand the role of the projection head in absorbing invariance, we evaluate methods on the embeddings instead of the predictor.

\begin{table*}[!h]
    \centering
    \caption{Results when evaluating on the representations or the embeddings.}
    \begin{tabular}{lccc}
        \toprule
        \multirow{2}{*}{Method} & \multicolumn{3}{c}{Rotation prediction ($R^2$)}   \\
        \cmidrule(lr){2-4} 
         & Representations & Embeddings & Change   \\ 
         \midrule
         VICRreg & 0.41 & 0.23 &\textcolor{red}{-0.18} \\
         SimCLR & 0.50 & 0.23 &\textcolor{red}{-0.28} \\
         Only Equivariance (Our predictor) & 0.60 & 0.39 & \textcolor{red}{-0.21}  \\
         EquiMod (Our predictor) & 0.60 & 0.39 & \textcolor{red}{-0.21}  \\
          SIE (Ours) &  0.73 & 0.60  & \textcolor{orange}{-0.13}   \\

         \bottomrule
    \end{tabular}
    \label{tab:quantitative-embs-reprs}
\end{table*}

As we can see in table~\ref{tab:quantitative-embs-reprs}, all methods suffer from a drop in performance after the projection head, highlighting the importance of the projection head's implicit equivariance. However the drop is significantly smaller for SIE, which achieves similar performance after the projection head to what other method achieve before it. This further help explain the quality of the predictor learned by SIE. We can notice that VICReg and SimCLR achieve a level of invariance after the projection head that is similar to SIE's before the projection head on the invariant part. As such the split of information was helpful to achieve invariance even before the projection head.

\begin{figure}[!th]
    \centering
    \includegraphics[width=\textwidth]{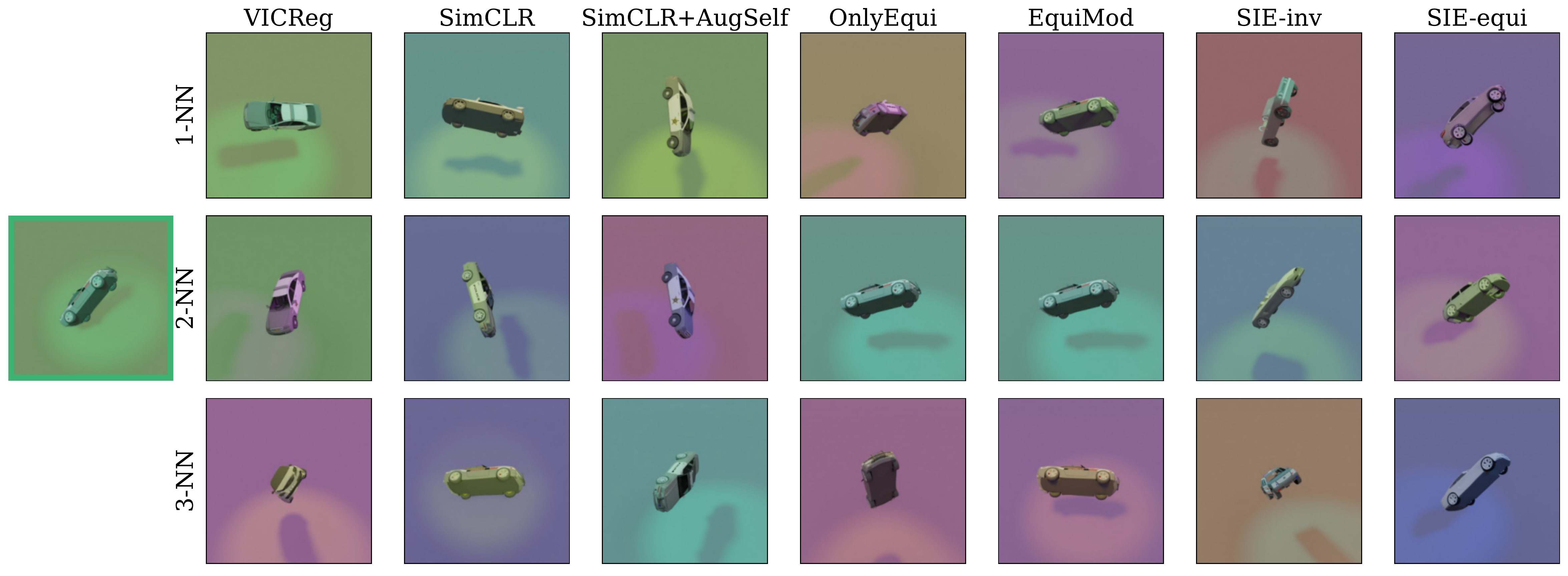}
    \caption{Retrieval of nearest embeddings. Starting from the representation associate to the object in the \textcolor{green}{green} frame on the left, we compute its nearest neighbours for all considered methods and show the 3 closest.}
    \label{fig:sample_embs}
\end{figure}

From a qualitative point of view, we reproduce figure~\ref{fig:sample_pred} which was done on the representations in figure~\ref{fig:sample_embs} which is done on the embeddings. We notice that SIE still achieves a similar level of equivariant performance, where the retrieved objects are in the same pose as the query. But for other methods the invariance is much stronger, and interestingly Only Equivariance does not appear to have much information about the object's pose. This further corroborates what we saw quantitatively.

\clearpage
\section{Predictor collapse to the identity\label{sec:pred-collapse}}

A phenomenon that motivated the introduction of the hypernetwork-based predictor is the collapse of certain predictors to the identity. This is something that can be observed in invariant self-supervised learning methods such as SimSiam~\cite{chen2020simsiam} where the predictor does not depend on the chosen images, but this can also happen in our framework. To illustrate this, we look at the weight matrix of the predictor when using a linear predictor where transformation parameters are concatenated to the embeddings, such as used by SEN or EquiMod.

\begin{figure}
    \centering
    \includegraphics[width=0.5\textwidth]{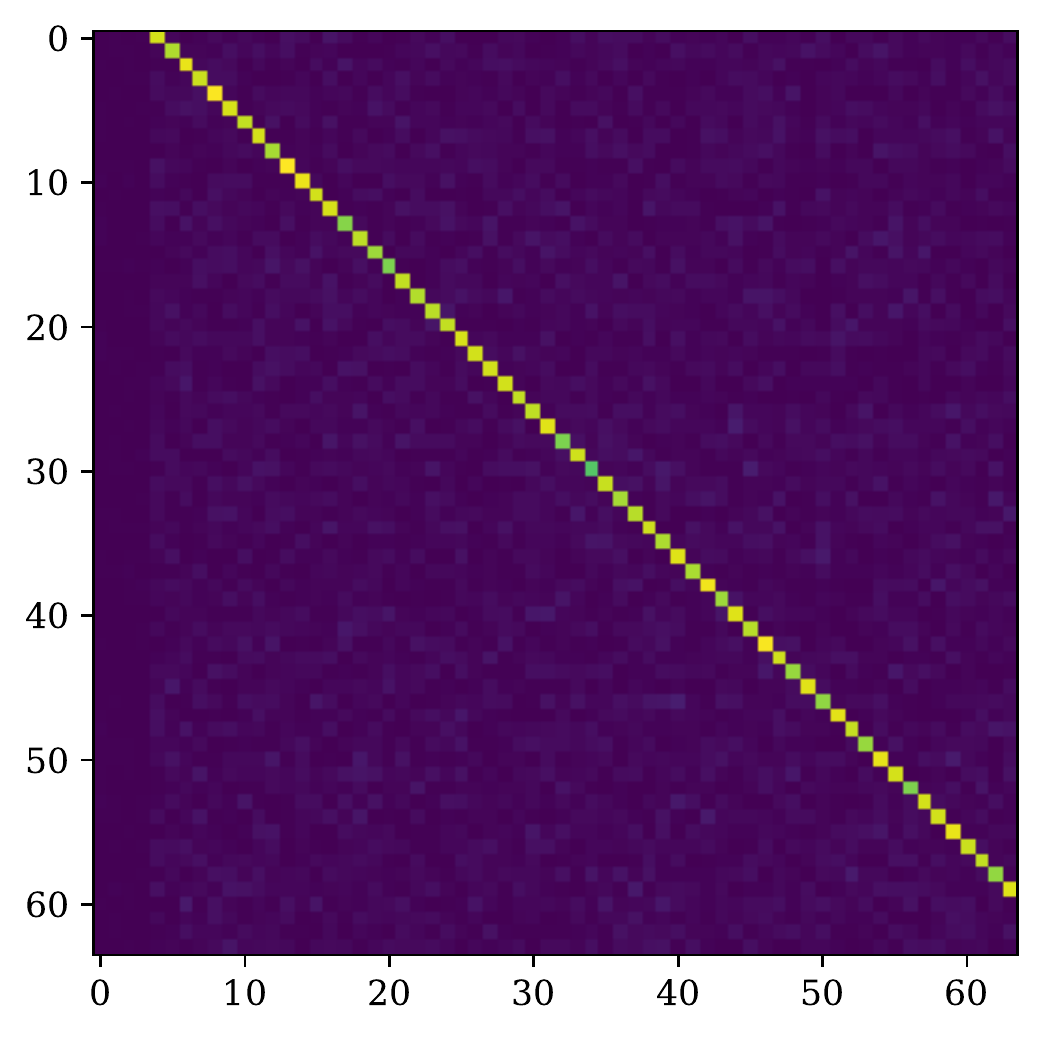}
    \caption{Collapse of the predictor to the identity. The first four columns which influence the rotation parameter are null and the rest of the predictor is close to the identity matrix.}
    \label{fig:id-pred}
\end{figure}

As we can see in figure~\ref{fig:id-pred}, the first four columns which are associated with the rotation are zero, and so the rotation parameters are ignored. The rest of the predictor is very close to the identity, which would suggest that the predictor effectively does nothing. As such the method collapsed to a classical invariant method in this case.

This highlights the need for a predictor where the information about the transformation cannot be ignored, such as our hypernetwork based design. This is a problem that must be taken into consideration when designing predictor based methods in complex scenarios.

\clearpage
\section{Results on rotation and color equivariance\label{sec:rot-col}}

While we previously trained models to be equivariant to rotation only, we study here the performance when trying to learn representations that are equivariant to both rotation and changes in color of the floor and light.

\begin{table*}[!h]
    \centering
    \caption{Quantitative evaluation of learned representations on invariant (classification) and equivariant (rotation prediction, color prediction) tasks. Equivariant methods are trained to be equivariant to rotation as well as floor and light hue. For each family of methods we highlight the best value in bold.$^ \dagger$ We train a supervised baseline for each evaluation.}
    \begin{tabular}{lccccccccc}
        \toprule
        Evaluation & \multicolumn{3}{c}{Classification (top-1)} & \multicolumn{3}{c}{Rotation prediction ($R^2$)} & \multicolumn{3}{c}{Color prediction ($R^2$)}   \\
        \cmidrule(lr){2-4} \cmidrule(lr){5-7}\cmidrule(lr){8-10}
         Representation part  (if applicable) & All & Inv. & Equi.  & All & Inv. & Equi. & All & Inv. & Equi. \\ 
         \midrule
         Supervised & 87.47 & & & 0.76 & & & 0.99 \\
         \midrule
         \textit{\textcolor{gray}{Invariant and parameter prediction methods}}\\
         VICReg & 84.74 & & &0.41 &   & & 0.06 \\
         VICReg, $g$ kept identical & 72.81  & & &0.56 &  & & 0.25\\
         SimCLR &\textbf{ 86.73} &  & & 0.50 &  & & 0.30 \\
         SimCLR, $g$ kept identical & 71.21  & & & 0.54 & & & 0.83 \\
          SimCLR + AugSelf & 85.34 & & & \textbf{0.75} & & & \textbf{0.98} \\
         \midrule
         \textit{\textcolor{gray}{Equivariant methods, \textbf{original predictor}}}\\
         Only Equivariance (Original predictor) & 86.70  & &  & 0.51  & &  & 0.26  \\
         Only Equivariance (Our predictor) & 85.25 & &   & 0.63 & &  & 0.97 \\
         EquiMod (Original predictor) & \textbf{86.93} & &  & 0.49 & &  & 0.91 \\
         EquiMod (Our predictor) & 86.48 & & & 0.58 & &  & 0.97 \\
          SIE (Ours)   & 80.93  & 80.60  & 77.28 & \textbf{0.67} & 0.43  & 0.68 & \textbf{0.98} & 0.14 & 0.98  \\

         \bottomrule
    \end{tabular}
    \label{tab:quantitative-rotcolor}
\end{table*}

As we can see in table~\ref{tab:quantitative-rotcolor}, all methods achieve a similar level of performance for classification and rotation prediction, although we can notice a slight drop for SIE. Looking at color prediction, we see that the performance is significantly increased from models trained to be equivariant only to rotation, achieving almost perfect performance on color prediction. The hypernetwork-based predictor also brings increased performance here, and the split representations bring a slight advantage in performance, though less noticeable than on rotation alone.

\clearpage
\section{Generalization to unseen rotations\label{sec:generalization}}

While we have trained methods on a certain set of rotations, we can wonder what happens when confronted with objects in unseen poses. To evaluate this, we generated rotations along the $z$-axis spanning the full circle, in increments of 5 degrees.

\begin{figure}[!h]
    \centering
    \includegraphics[width=0.4\textwidth]{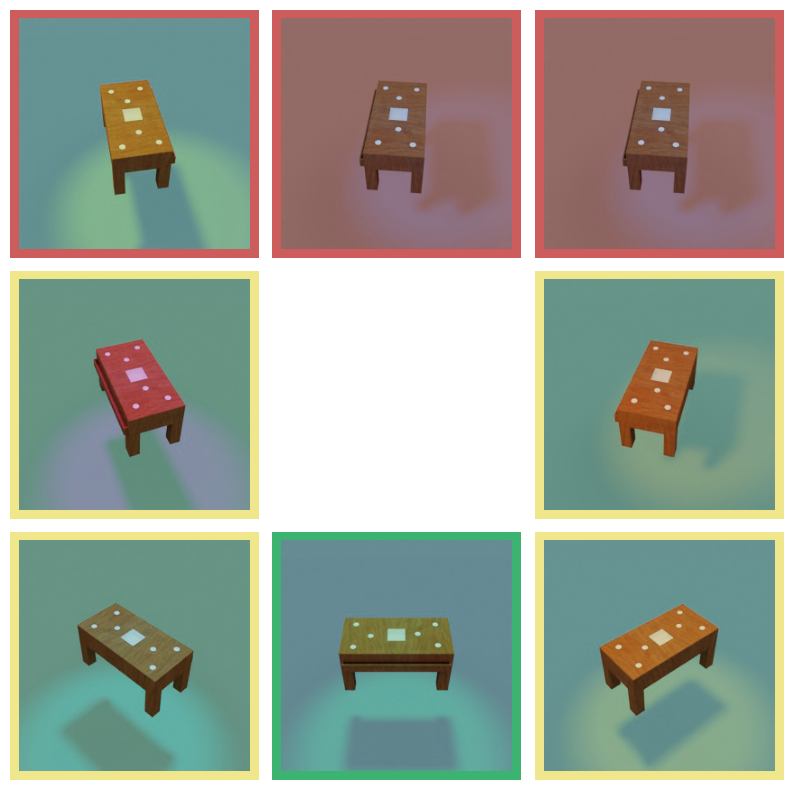}
    \includegraphics[width=0.4\textwidth]{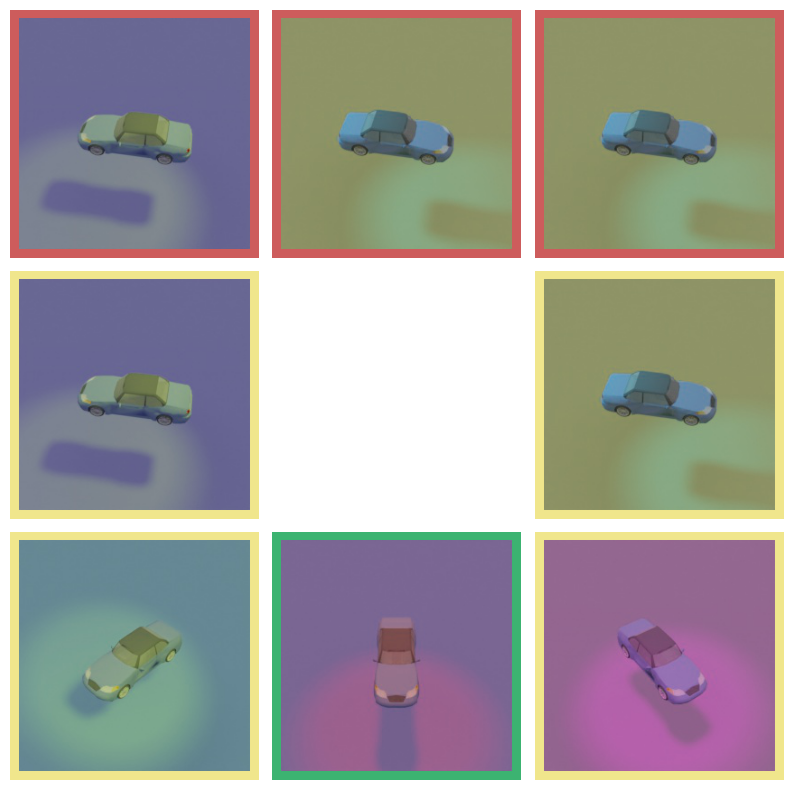}
    \caption{Generalization to unseen rotations during training. Starting from the canonical view (\textcolor{green}{green} frame), we apply rotations through the predictor of a trained SIE. Rotation were either possibly seen during training (\textcolor{yellow}{yellow} frame) or could not have been seen (\textcolor{red}{red} frame).}
    \label{fig:generalization}
\end{figure}

As we can see in figure~\ref{fig:generalization}, while the model manages to extrapolate reasonably well for the table, it fails to do so for the car, giving retrieved images that are at the maximum rotation that could have been seen during training. There may be multiple reasons for this. This could come from the encoder that isn't able to encode these rotated images properly, but this could also come from the predictor that is not able to learn a transformation for these unseen angles. Most likely, this phenomenon is a combination of the two.

\clearpage
\section{Dataset generation\label{sec:dataset}}

\subsection{Detailed generation process}
\begin{table}[]
    \centering
    \caption{Values of the factors of variation used for the generation of 3DIEBench. Each value is samples unofrmly from the given interval. Object rotation is generated as Tayt-Bryan angles using extrinsic rotations. Light position is expressed in spherical coordinates. }
    \begin{tabular}{lcc}
    \toprule
    Parameter & Minimum value & Maximum value\\
    \midrule
     Object rotation X & $-\frac{\pi}{2}$  & $\frac{\pi}{2}$ \\
     Object rotation Y & $-\frac{\pi}{2}$  & $\frac{\pi}{2}$ \\
     Object rotation Z & $-\frac{\pi}{2}$ & $\frac{\pi}{2}$\\
     Floor hue & 0 & 1 \\
     Light hue & 0 & 1 \\
     Light $\theta$ & $0$ & $\frac{\pi}{4}$\\
     Light $\phi$& $0$ & $2\pi$ \\
    \bottomrule
    \end{tabular}
    \label{tab:data_gen_values}
\end{table}

The basis of 3DIEBench is the subset of ShapeNetCore coming from 3D Warehouse. This gives 52462 models spanning 55 classes. We split the dataset into a training and validation part, containing respectively 80\% and 20\% of the objects.
Starting from a given 3D model, we generate 50 different scenes by changing factors of variation using the ranges described in table~\ref{tab:data_gen_values}. We change the light position to ensure that the objects shadow is not informative and cannot be used as a crutch by the network. The range of rotations is limited in order to make the problem more tractable. As we see in the results, it is still very challenging for existing approaches.

The generation of all images takes around 500 hours on a single NVIDIA V100 GPU, but can be easily parallelized.

\subsection{Additional data samples}

\begin{figure*}[!th]
    \centering
    \includegraphics[width=\textwidth]{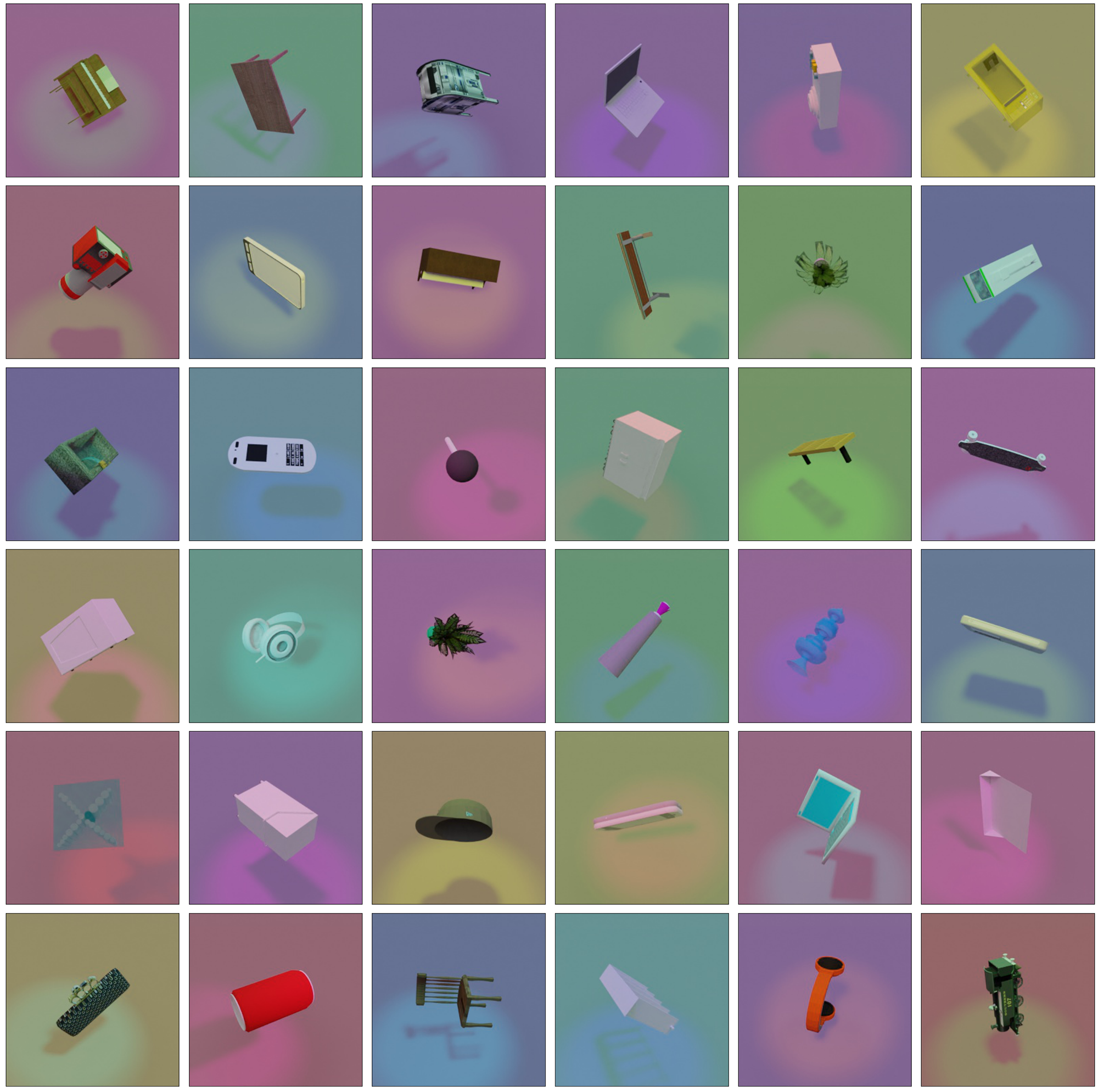}
    \caption{Samples from 3DIEBench}
    \label{fig:sample_1}
\end{figure*}
\begin{figure*}[!th]
    \centering
    \includegraphics[width=\textwidth]{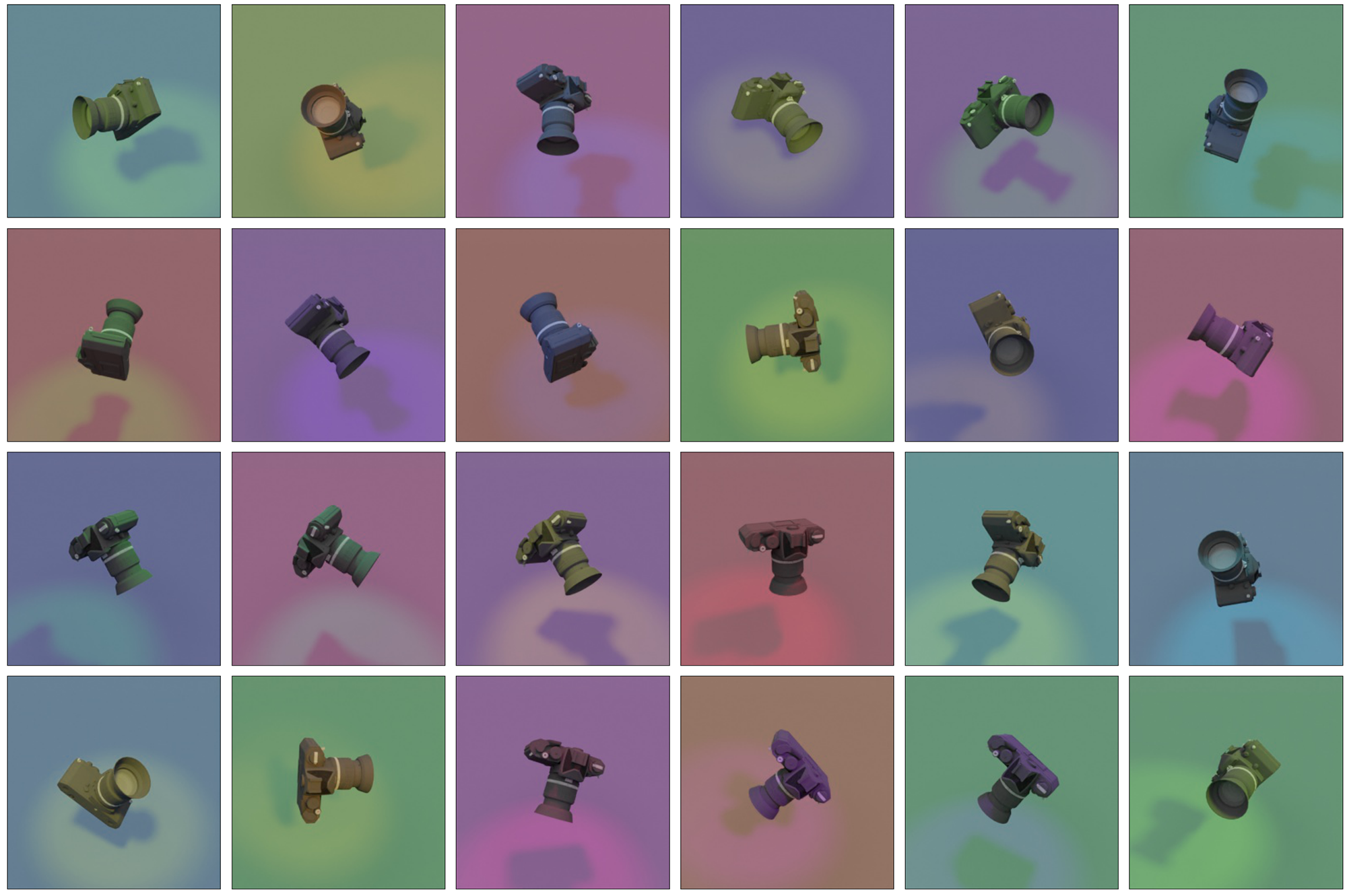}
    \caption{Samples from one object from 3DIEBench}
    \label{fig:sample_2}
\end{figure*}
\begin{figure*}[!th]
    \centering
    \includegraphics[width=\textwidth]{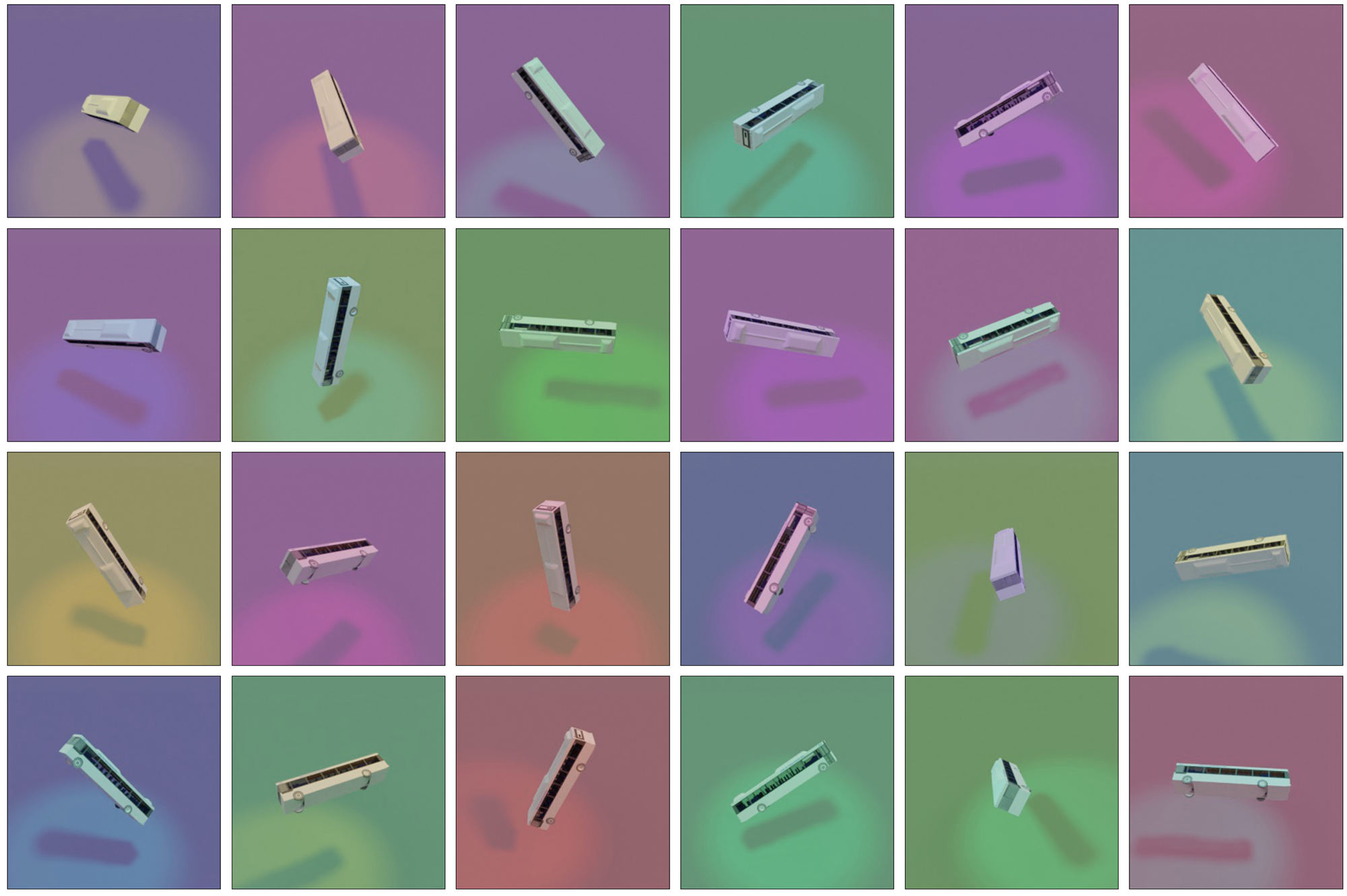}
    \caption{Samples from one object from 3DIEBench}
    \label{fig:sample_3}
\end{figure*}

\end{document}